%% file: main.tex
\pdfoutput=1
\documentclass[11pt]{article}
\usepackage{eacl2017}
\usepackage{times}
\usepackage{url}
\usepackage{latexsym}

\eaclfinalcopy 

\usepackage{xspace}
\usepackage{mathtools}
\usepackage{amsmath}
\usepackage{amssymb}
\usepackage[usenames,dvipsnames]{color}
\usepackage{multirow}
\usepackage{tikz}
\usepackage{pifont}
\usepackage[inline]{enumitem}
\usetikzlibrary{positioning}
\usetikzlibrary{calc}
\usetikzlibrary{fit}
\usetikzlibrary{decorations.text}
\usetikzlibrary{shapes.misc}
\usetikzlibrary{shapes.geometric,arrows}
\usepackage{colortbl}

\newcommand{\eg}{e.g.\xspace}
\newcommand{\ie}{i.e.\xspace}

\newcommand{\system}[1]{\texttt{#1}\xspace}

\newcommand{\secref}[1]{Section~\ref{#1}\xspace}
\newcommand{\tabref}[2][]{Table#1~\ref{#2}\xspace}
\newcommand{\figref}[2][]{Figure#1~\ref{#2}\xspace}

\newcommand{\equref}[1]{Equation~(\ref{#1})\xspace}

\newcommand{\bb}[1]{\mathbb{#1}}
\newcommand{\R}{\bb{R}}
\newcommand{\mat}[2][]{\boldsymbol{#2}_{#1}}

\renewcommand{\vec}[2][]{\boldsymbol{#2}^{#1}}

\newcommand{\softmax}{\textrm{softmax}}

\newcommand{\T}{\mathstrut\scriptscriptstyle\top}

\newcommand{\memnn}{\system{MemN2N}}
\newcommand{\gmemnn}{\system{GMemN2N}}

\newcommand{\babi}{\system{bAbI}}
\newcommand{\dialogbabi}{\system{Dialog bAbI}}
\newcommand{\dstc}{\system{DSTC-2}}

\newcommand{\RNum}[1]{\uppercase\expandafter{\romannumeral #1\relax}}

\newcommand{\myparagraph}[1]{\vspace{0.6\baselineskip}\noindent{\textbf{#1}}~}

\newcommand{\specialcell}[2][c]{%
  \begin{tabular}[#1]{@{}l@{}}#2\end{tabular}}

\title{Gated End-to-End Memory Networks}

\author{Fei Liu \thanks{work done as an Intern at Xerox Research Centre Europe}\\
  The University of Melbourne \\
  Victoria, Australia \\
  {\tt fliu3@student.unimelb.edu.au} \\\And
  Julien Perez \\
  Xerox Research Centre Europe \\
  Grenoble, France \\
  {\tt julien.perez@xrce.xerox.com} \\}

\date{}

\begin{document}
\maketitle
\begin{abstract}
Machine reading using differentiable reasoning models has recently shown remarkable progress. In this context, End-to-End trainable Memory Networks (\memnn) have demonstrated promising performance on simple natural language based reasoning tasks such as factual reasoning and basic deduction. However, other tasks, namely multi-fact question-answering, positional reasoning or dialog related tasks, remain challenging particularly due to the necessity of more complex interactions between the memory and controller modules composing this family of models. In this paper, we introduce a novel end-to-end memory access regulation mechanism inspired by the current progress on the connection short-cutting principle in the field of computer vision. Concretely, we develop a Gated End-to-End trainable Memory Network architecture (\gmemnn). From the machine learning perspective, this new capability is learned in an end-to-end fashion without the use of any additional supervision signal which is, as far as our knowledge goes, the first of its kind. Our experiments show significant improvements on the most challenging tasks in the 20 \babi dataset, without the use of any domain knowledge. Then, we show improvements on the \dialogbabi tasks including the real human-bot conversion-based Dialog State Tracking Challenge (\dstc) dataset. On these two datasets, our model sets the new state of the art.
\end{abstract}

\input{intro}
\input{bg}
\input{model}
\input{experiments}
\input{conc}

\bibliographystyle{eacl2017}
\bibliography{ml}

\end{document}

%% file: intro.tex
\section{Introduction}
\label{sec:intro}

Deeper Neural Network models are more difficult to train and recurrency tends to complexify this optimization problem \cite{Srivastava+:2015}. While Deep Neural Network architectures have shown superior performance in numerous areas, such as image, speech recognition and more recently text, the complexity of optimizing such large and non-convex parameter sets remains a challenge. Indeed, the so-called vanishing/exploding gradient problem has been mainly addressed using:
\begin{enumerate*}
\item algorithmical responses, \eg, normalized initialization stategies \cite{LeCun98,GlorotB10};
\item architectural ones, \eg, intermediate normalization layers which facilitate the convergence of networks composed of tens of hidden layers \cite{HeZRS15,SaxeMG13}.
\end{enumerate*}
Another problem of memory-enhanced neural models is the necessity of regulating memory access at the controller level. Memory access operations can be supervised \cite{KumarIOIBGZPS16} and the number of times they are performed tends to be fixed apriori \cite{Sukhbaatar+:2015}, a design choice which tends to be based on the presumed degree of difficulty of the task in question. Inspired by the recent success of object recognition in the field of computer vision \cite{SrivastavaGS15,Srivastava+:2015}, we investigate the use of a {\it gating} mechanism in the context of End-to-End Memory Networks (\memnn) \cite{Sukhbaatar+:2015} in order to regulate the access to the memory blocks in a differentiable fashion. The formulation is realized by gated connections between the memory access layers and the controller stack of a \memnn. As a result, the model is able to dynamically determine how and when to skip its memory-based reasoning process.

\myparagraph{Roadmap:} \secref{sec:bg} reviews state-of-the-art Memory Network models, connection short-cutting in neural networks and memory dynamics. In \secref{sec:model}, we propose a differentiable gating mechanism in \memnn. \secref{sec:experimentsqa} and \ref{sec:experimentsdialog} present a set of experiments on the 20 \babi reasoning tasks and the \dialogbabi dataset. We report new state-of-the-art results on several of the most challenging tasks of the set, namely positional reasoning, $3$-argument relation and the \dstc task while maintaining equally competitive results on the rest.

%% file: bg.tex
\section{Related Work}
\label{sec:bg}

This section starts with an introduction of the primary elements of \memnn. Then, we review two key elements relevant to this work, namely shortcut connections in neural networks in and memory dynamics in such models.

\subsection{End-to-End Memory Networks}

The \memnn architecture, introduced by \newcite{Sukhbaatar+:2015}, consists of two main components: supporting memories and final answer prediction. Supporting memories are in turn comprised of a set of input and output memory representations with memory cells. The input and output memory cells, denoted by $\vec{m}_{i}$ and $\vec{c}_{i}$, are obtained by transforming the input context $x_{1},\ldots,x_{n}$ (or stories) using two embedding matrices $\mat{A}$ and $\mat{C}$ (both of size $d \times |V|$ where $d$ is the embedding size and $|V|$ the vocabulary size) such that $\vec{m}_{i} = \mat{A}\Phi(x_{i})$ and $\vec{c}_{i} = \mat{C}\Phi(x_{i})$ where $\Phi(\cdot)$ is a function that maps the input into a bag of dimension $|V|$. Similarly, the question $q$ is encoded using another embedding matrix $\mat{B} \in \R^{d \times |V|}$, resulting in a question embedding $\vec{u} = \mat{B}\Phi(q)$. The input memories $\{\vec{m}_{i}\}$, together with the embedding of the question $\vec{u}$, are utilized to determine the relevance of each of the stories in the context, yielding a vector of attention weights 
\begin{equation}
p_{i} = \softmax(\vec[\T]{u}\vec{m}_{i})
\end{equation}
where $\softmax(a_{i}) = \dfrac{e^{a_{i}}}{\sum_{j \in [1, n]}e^{a_{j}}}$. Subsequently, the response $\vec{o}$ from the output memory is constructed by the weighted sum: 
\begin{equation}
\vec{o} = \sum_{i}p_{i}\vec{c}_{i}
\end{equation}

For more difficult tasks requiring multiple supporting memories, the model can be extended to include more than one set of input/output memories by stacking a number of memory layers. In this setting, each memory layer is named a hop and the $(k+1)^{\textrm{th}}$ hop takes as input the output of the $k^{\textrm{th}}$ hop:
\begin{equation}
\label{equ:memnn}
\vec[k+1]{u} = \vec[k]{o} + \vec[k]{u}
\end{equation}

Lastly, the final step, the prediction of the answer to the question $q$, is performed by 
\begin{equation}
\label{equ:finalpred}
\hat{\vec{a}} = \softmax(\mat{W}(\vec[K]{o} + \vec[K]{u}))
\end{equation}
where $\hat{\vec{a}}$ is the predicted answer distribution, $\mat{W} \in \R^{|V| \times d}$ is a parameter matrix for the model to learn and $K$ the total number of hops.

\subsection{Shortcut Connections} 

Shortcut connections have been studied from both the theoretical and practical point of view in the general context of neural network architectures \cite{Bishop1995,ripley96pattern}. More recently Residual Networks \cite{HeZRS15a} and Highway Networks \cite{SrivastavaGS15,Srivastava+:2015} have been almost simultaneously proposed. While the former utilizes a residual calculus, the latter formulates a differentiable gateway mechanism as proposed in Long-Short Terms Memory Networks in order to cope with long-term dependency issues in the dataset in an end-to-end trainable manner. These two mechanisms were proposed as a structural solution to the so-called vanishing gradient problem by allowing the model to shortcut its layered transformation structure when necessary.

\subsection{Memory Dynamics} 

The necessity of dynamically regulating the interaction between the so-called controller and the memory blocks of a Memory Network model has been study in \cite{KumarIOIBGZPS16,XiongMS16}. In these works, the number of exchanges between the controller stack and the memory module of the network is either monitored in a hard supervised manner in the former or fixed apriori in the latter. 

In this paper, we propose an end-to-end supervised model, with an automatically learned gating mechanism, to perform dynamic regulation of memory interaction. The next section presents the formulation of this new Gated End-to-End Memory Networks (\gmemnn). This contribution can be placed in parallel to the recent transition from Memory Networks with hard attention mechanism \cite{WestonCB14} to \memnn with attention values obtained by a $\softmax$ function and end-to-end supervised \cite{Sukhbaatar+:2015}.

%% file: model.tex
\section{Gated End-to-End Memory Network}
\label{sec:model}

In this section, the elements behind residual learning and highway neural models are given. Then, we introduce the proposed model of memory access gating in a \memnn.

\subsection{Highway and Residual Networks}

Highway Networks, first introduced by \newcite{SrivastavaGS15}, include a transform gate $\textrm{T}$ and a carry gate $\textrm{C}$, allowing the network to learn how much information it should transform or carry to form the input to the next layer. Suppose the original network is a plain feed-forward neural network:
\begin{equation}
\vec{y} = \textrm{H}(\vec{x})
\end{equation}
where $\textrm{H}(\vec{x})$ is a non-linear transformation of its input $\vec{x}$. The generic form of Highway Networks is formulated as:
\begin{equation}
\vec{y} = \textrm{H}(\vec{x})\odot\textrm{T}(\vec{x}) + \vec{x}\odot\textrm{C}(\vec{x})
\end{equation}
where the transform and carry gates, $\textrm{T}(\vec{x})$ and $\textrm{C}(\vec{x})$, are defined as non-linear transformation functions of the input $\vec{x}$ and $\odot$ the Hadamard product. As suggested in \cite{SrivastavaGS15,Srivastava+:2015}, we choose to focus, in the following of this paper, on a simplified version of Highway Networks where the carry gate is replaced by $1 - \textrm{T}(\vec{x})$:
\begin{equation}
\vec{y} = \textrm{H}(\vec{x})\odot\textrm{T}(\vec{x}) + \vec{x}\odot(1 - \textrm{T}(\vec{x}))
\end{equation}
where $\textrm{T}(x) = \sigma(\mat[T]{W}\vec{x} + \vec{b}_T)$ and $\sigma$ is the sigmoid function. In fact, Residual Networks can be viewed as a special case of Highway Networks where both the transform and carry gates are substituted by the identity mapping function:
\begin{align}
\vec{y} &= \textrm{H}(\vec{x}) + \vec{x}
\end{align}
thereby forming a hard-wired shortcut connection $\vec{x}$.

\subsection{Gated End-to-End Memory Networks}

\begin{figure*}[tb]
\begin{center}
\resizebox{.9\textwidth}{!}{
\input{resmemn2n.tex}
}
\end{center}
\caption{\label{fig:resmemn2n}Illustration of the proposed \gmemnn model with $3$ hops.}
\end{figure*}
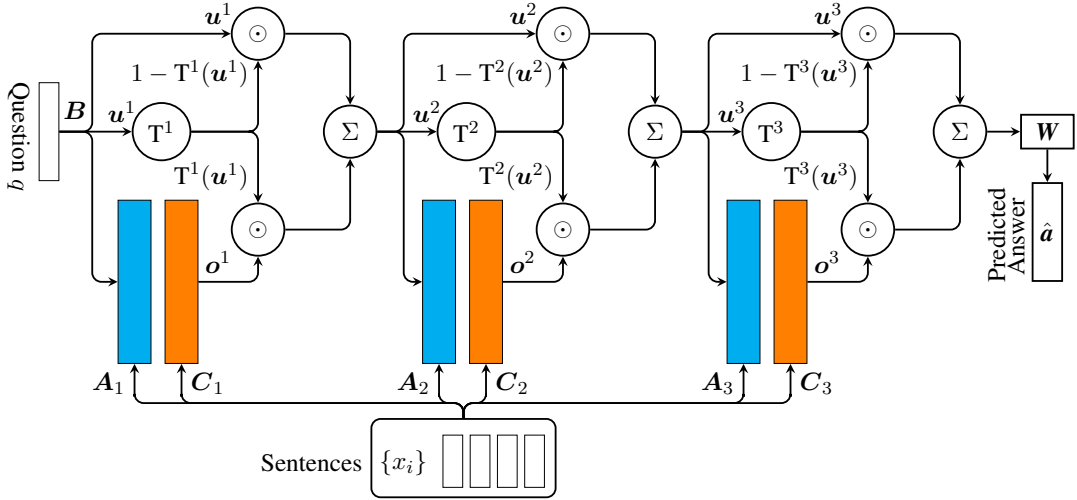

Arguably, \equref{equ:memnn} can be considered as a form of residuality with $\vec[k]{o}$ working as the residual function and $\vec[k]{u}$ the shortcut connection. However, as discussed in \cite{Srivastava+:2015}, in contrast to the hard-wired skip connection in Residual Networks, one of the advantages of Highway Networks is the adaptive gating mechanism, capable of learning to dynamically control the information flow based on the current input. Therefore, we adopt the idea of the adaptive gating mechanism of Highway Networks and integrate it into \memnn. The resulting model, named \textit{Gated End-to-End Memory Networks} (\gmemnn) and illustrated in \figref{fig:resmemn2n}, is capable of dynamically conditioning the memory reading operation on the controller state $\vec[k]{u}$ at each hop. Concretely, we reformulate \equref{equ:memnn} into:
\begin{align}
\textrm{T}^{k}(\vec[k]{u}) &= \sigma(\mat[T]{W}^{k}\vec[k]{u} + \vec[k]{b}_T) \\
\vec[k+1]{u}  &= \vec[k]{o}\odot\textrm{T}^{k}(\vec[k]{u}) + \vec[k]{u}\odot(1 - \textrm{T}^{k}(\vec[k]{u}))
\end{align}
where $\mat[T]{W}^{k}$ and $\vec[k]{b}$ are the hop-specific parameter matrix and bias term for the $k^{\textrm{th}}$ hop and $\textrm{T}^{k}(x)$ the transform gate for the $k^{\textrm{th}}$ hop. Similar to the two weight tying schemes of the embedding matrices introduced in \cite{Sukhbaatar+:2015}, we also explore two types of constraints on $\mat[T]{W}^{k}$ and $\vec[k]{b}_{T}$:
\begin{enumerate}[noitemsep,topsep=0pt]
	\itemsep0em
	\item \textbf{Global:} all the weight matrices $\mat[T]{W}^{k}$ and bias terms $\vec[k]{b}_{T}$ are shared across different hops, \ie, $\mat[T]{W}^{1} = \mat[T]{W}^{2} = \ldots = \mat[T]{W}^{K}$ and $\vec[1]{b}_{T} = \vec[2]{b}_{T} = \ldots = \vec[K]{b}_{T}$.
	\item \textbf{Hop-specific:} each hop has its specific weight matrix $\mat[T]{W}^{k}$ and bias term $\vec[k]{b}_{T}$ for $k \in [1, K]$ and they are optimized independently.
\end{enumerate}

%% file: resmemn2n.tex
\begin{tikzpicture}

\node[draw,thick,align=center,minimum height=1.2cm,minimum width=2.8cm,rounded corners] at (7.0,0.5) (stories) {};
\node[anchor=north west,shift={(0mm,2.5mm)}] (stories_xi) at (stories.west){$\{x_i\}$};
\node[draw,align=center,minimum height=0.8cm,minimum width=0.3cm] [right=0.1of stories_xi,shift={(0mm,0.3mm)}] (stories_sent1) {};
\node[draw,align=center,minimum height=0.8cm,minimum width=0.3cm] [right=0.1of stories_sent1] (stories_sent2) {};
\node[draw,align=center,minimum height=0.8cm,minimum width=0.3cm] [right=0.1of stories_sent2] (stories_sent3) {};
\node[draw,align=center,minimum height=0.8cm,minimum width=0.3cm] [right=0.1of stories_sent3] (stories_sent4) {};
\node[anchor=north west,shift={(-18mm,2mm)}] (sentences_label) at (stories.west){Sentences};

\node[draw,align=center,minimum height=1.5cm,minimum width=0.3cm] at (0.7,5.5) (question) {};
\node[anchor=north west,shift={(0mm,9mm)},rotate=-90] (question_label) at (question.west){Question $q$};

\node[draw,align=center,minimum height=2.5cm,minimum width=0.5cm,fill=cyan] at (2.0,3.2) (A1) {};
\node[draw,align=center,minimum height=2.5cm,minimum width=0.5cm,fill=orange] [right=0.2of A1] (C1) {};

\node[draw,thick,align=center,circle,minimum size=0.8cm] [right=1.1 of question] (Tx1) {$\textrm{T}^{1}$};

\node[draw,thick,align=center,circle,minimum size=0.8cm] [right=0.6 of Tx1,shift={(0mm,15mm)}] (Prod11) {$\odot$};
\node[draw,thick,align=center,circle,minimum size=0.8cm] [right=0.6 of Tx1,shift={(0mm,-15mm)}] (Prod12) {$\odot$};

\node[draw,thick,align=center,circle,minimum size=0.8cm] [right=2.0 of Tx1] (Sum1) {$\Sigma$};

\draw [rounded corners,thick,->,>=stealth] (question) -- ($(question.east) + (0.5,0.0)$) node [midway,above] (B_label) {$\mat{B}$} |- (A1);
\draw [rounded corners,thick,->,>=stealth] (stories) |- ($(A1.south) + (0.0,-0.6)$) -| (A1.south) node [left,shift={(0mm,-3mm)}] (A1_label) {$\mat[1]{A}$};
\draw [rounded corners,thick,->,>=stealth] (stories) |- ($(C1.south) + (0.0,-0.6)$) -| (C1.south) node [right,shift={(0mm,-3mm)}] (C1_label) {$\mat[1]{C}$};

\draw [rounded corners,thick,->,>=stealth] (question) -- (Tx1) node [above,shift={(-6mm,0mm)}] (u1_label) {$\vec[1]{u}$};
\draw [rounded corners,thick,->,>=stealth] (question) -- ($(question.east) + (0.5,0.0)$) |- (Prod11) node [above,shift={(-6mm,0mm)}] (u1_label) {$\vec[1]{u}$};

\draw [rounded corners,thick,->,>=stealth] (Tx1) -- ($(Tx1.east) + (0.8,0.0)$) -| (Prod11.south) node [left,shift={(0mm,-2mm)}] () {$1 - \textrm{T}^{1}(\vec[1]{u})$};
\draw [rounded corners,thick,->,>=stealth] (Tx1) -- ($(Tx1.east) + (0.8,0.0)$) -| (Prod12.north) node [left,shift={(0mm,4mm)}] () {$\textrm{T}^{1}(\vec[1]{u})$};

\draw [rounded corners,thick,->,>=stealth] (C1.east) -- ($(C1.east) + (0.3,0.0)$)  node [above,shift={(0mm,0mm)}] () {$\vec[1]{o}$} -| (Prod12.south);

\draw [rounded corners,thick,->,>=stealth] (Prod11.east) -- ($(Prod11.east) + (0.3,0.0)$) -| (Sum1.north);
\draw [rounded corners,thick,->,>=stealth] (Prod12.east) -- ($(Prod12.east) + (0.3,0.0)$) -| (Sum1.south);

\node[draw,align=center,minimum height=2.5cm,minimum width=0.5cm,fill=cyan] [right=3.4 of C1] (A2) {};
\node[draw,align=center,minimum height=2.5cm,minimum width=0.5cm,fill=orange] [right=0.2of A2] (C2) {};

\node[draw,thick,align=center,circle,minimum size=0.8cm] [right=0.9 of Sum1] (Tx2) {$\textrm{T}^{2}$};

\node[draw,thick,align=center,circle,minimum size=0.8cm] [right=0.6 of Tx2,shift={(0mm,15mm)}] (Prod21) {$\odot$};
\node[draw,thick,align=center,circle,minimum size=0.8cm] [right=0.6 of Tx2,shift={(0mm,-15mm)}] (Prod22) {$\odot$};

\node[draw,thick,align=center,circle,minimum size=0.8cm] [right=2.0 of Tx2] (Sum2) {$\Sigma$};

\draw [rounded corners,thick,->,>=stealth] (Sum1) -- ($(Sum1.east) + (0.4,0.0)$) |- (A2);
\draw [rounded corners,thick,->,>=stealth] (stories) |- ($(A2.south) + (0.0,-0.6)$) -| (A2.south) node [left,shift={(0mm,-3mm)}] (A2_label) {$\mat[2]{A}$};
\draw [rounded corners,thick,->,>=stealth] (stories) |- ($(C2.south) + (0.0,-0.6)$) -| (C2.south) node [right,shift={(0mm,-3mm)}] (C2_label) {$\mat[2]{C}$};

\draw [rounded corners,thick,->,>=stealth] (Sum1) -- (Tx2) node [above,shift={(-6mm,0mm)}] (u2_label) {$\vec[2]{u}$};
\draw [rounded corners,thick,->,>=stealth] (Sum1) -- ($(Sum1.east) + (0.4,0.0)$) |- (Prod21) node [above,shift={(-6mm,0mm)}] (u2_label) {$\vec[2]{u}$};

\draw [rounded corners,thick,->,>=stealth] (Tx2) -- ($(Tx2.east) + (0.8,0.0)$) -| (Prod21.south) node [left,shift={(0mm,-2mm)}] () {$1 - \textrm{T}^{2}(\vec[2]{u})$};
\draw [rounded corners,thick,->,>=stealth] (Tx2) -- ($(Tx2.east) + (0.8,0.0)$) -| (Prod22.north) node [left,shift={(0mm,4mm)}] () {$\textrm{T}^{2}(\vec[2]{u})$};

\draw [rounded corners,thick,->,>=stealth] (C2.east) -- ($(C2.east) + (0.3,0.0)$)  node [above,shift={(0mm,0mm)}] () {$\vec[2]{o}$} -| (Prod22.south);

\draw [rounded corners,thick,->,>=stealth] (Prod21.east) -- ($(Prod21.east) + (0.3,0.0)$) -| (Sum2.north);
\draw [rounded corners,thick,->,>=stealth] (Prod22.east) -- ($(Prod22.east) + (0.3,0.0)$) -| (Sum2.south);

\node[draw,align=center,minimum height=2.5cm,minimum width=0.5cm,fill=cyan] [right=3.4 of C2] (A3) {};
\node[draw,align=center,minimum height=2.5cm,minimum width=0.5cm,fill=orange] [right=0.2of A3] (C3) {};

\node[draw,thick,align=center,circle,minimum size=0.8cm] [right=0.9 of Sum2] (Tx3) {$\textrm{T}^{3}$};

\node[draw,thick,align=center,circle,minimum size=0.8cm] [right=0.6 of Tx3,shift={(0mm,15mm)}] (Prod31) {$\odot$};
\node[draw,thick,align=center,circle,minimum size=0.8cm] [right=0.6 of Tx3,shift={(0mm,-15mm)}] (Prod32) {$\odot$};

\node[draw,thick,align=center,circle,minimum size=0.8cm] [right=2.0 of Tx3] (Sum3) {$\Sigma$};

\draw [rounded corners,thick,->,>=stealth] (Sum2) -- ($(Sum2.east) + (0.4,0.0)$) |- (A3);
\draw [rounded corners,thick,->,>=stealth] (stories) |- ($(A3.south) + (0.0,-0.6)$) -| (A3.south) node [left,shift={(0mm,-3mm)}] (A3_label) {$\mat[3]{A}$};
\draw [rounded corners,thick,->,>=stealth] (stories) |- ($(C3.south) + (0.0,-0.6)$) -| (C3.south) node [right,shift={(0mm,-3mm)}] (C3_label) {$\mat[3]{C}$};

\draw [rounded corners,thick,->,>=stealth] (Sum2) -- (Tx3) node [above,shift={(-6mm,0mm)}] (u3_label) {$\vec[3]{u}$};
\draw [rounded corners,thick,->,>=stealth] (Sum2) -- ($(Sum2.east) + (0.4,0.0)$) |- (Prod31) node [above,shift={(-6mm,0mm)}] (u3_label) {$\vec[3]{u}$};

\draw [rounded corners,thick,->,>=stealth] (Tx3) -- ($(Tx3.east) + (0.8,0.0)$) -| (Prod31.south) node [left,shift={(0mm,-2mm)}] () {$1 - \textrm{T}^{3}(\vec[3]{u})$};
\draw [rounded corners,thick,->,>=stealth] (Tx3) -- ($(Tx3.east) + (0.8,0.0)$) -| (Prod32.north) node [left,shift={(0mm,4mm)}] () {$\textrm{T}^{3}(\vec[3]{u})$};

\draw [rounded corners,thick,->,>=stealth] (C3.east) -- ($(C3.east) + (0.3,0.0)$)  node [above,shift={(0mm,0mm)}] () {$\vec[3]{o}$} -| (Prod32.south);

\draw [rounded corners,thick,->,>=stealth] (Prod31.east) -- ($(Prod31.east) + (0.3,0.0)$) -| (Sum3.north);
\draw [rounded corners,thick,->,>=stealth] (Prod32.east) -- ($(Prod32.east) + (0.3,0.0)$) -| (Sum3.south);

\node[draw,thick,align=center,minimum width=0.8cm] [right=0.5 of Sum3] (W) {$\textbf{\textit{W}}$};

\node[draw,thick,align=center,minimum height=1.5cm] [below=0.5 of W] (a) {$\hat{\textbf{\textit{a}}}$};
\node[anchor=north west,shift={(-8mm,-9mm)},rotate=90] (answer_label) at (a.west){Predicted};
\node[anchor=north west,shift={(-5mm,-8mm)},rotate=90] (answer_label) at (a.west){Answer};

\draw [rounded corners,thick,->,>=stealth] (Sum3.east) -- (W.west);
\draw [rounded corners,thick,->,>=stealth] (W.south) -- (a.north);

\end{tikzpicture}

%% file: experiments.tex
\section{QA bAbI Experiments}
\label{sec:experimentsqa}
In this section, we first describe the natural language reasoning dataset we use in our experiments. Then, the experimental setup is detailed. Lastly, we present the results and analyses.

\subsection{Dataset and Data Preprocessing}

The 20 \babi tasks \cite{WestonBCM15} have been employed for the experiments (using v1.2 of the dataset). In this dataset, a given QA task consists of a set of statements, followed by a question whose answer is typically a single word (in a few tasks, answers are a set of words). The answer is available to the model at training time but must be predicted at test time. The dataset consists of 20 different tasks with various emphases on different forms of reasoning. For each question, only a certain subset of the statements contains information needed for the answer, and the rest are essentially irrelevant distractors. As in \cite{Sukhbaatar+:2015}, our model is fully end-to-end trained without any additional supervision other than the answers themselves. Formally, for one of the 20 QA tasks, we are given example problems, each having a set of $I$ sentences $\{x_i\}$ (where $I \leq 320$), a question sentence $q$ and answer $a$. Let the $j^{\textrm{th}}$ word of sentence $i$ be $x_{ij}$, represented by a one-hot vector of length $|V|$. The same representation is used for the question $q$ and answer $a$. Two versions of the data are used, one that has 1,000 training problems per task and the other with 10,000 per task.

\subsection{Training Details}
\label{sec:trainingdetails}

As suggested in \cite{Sukhbaatar+:2015}, $10\%$ of the \babi training set was held-out to form a validation set for hyperparameter tuning. Moreover, we use the so-called position encoding, adjacent weight tying, and temporal encoding with $10\%$ random noise. Learning rate $\eta$ is initially assigned a value of $0.005$ with exponential decay applied every $25$ epochs by $\eta/2$ until $100$ epochs are reached. Linear start is used in all our experiments as proposed by \newcite{Sukhbaatar+:2015}. With linear start, the $\softmax$ in each memory layer is removed and re-inserted after $20$ epochs. Batch size is set to $32$ and gradients with an $\ell_2$ norm larger than $40$ are divided by a scalar to have norm $40$. All weights are initialized randomly from a Gaussian distribution with zero mean and $\sigma = 0.1$ except for the transform gate bias $\vec[k]{b}_{T}$ which we empirically set the mean to $0.5$. Only the most recent $50$ sentences are fed into the model as the memory and the number of memory hops is $3$. In all our experiments, we use the embedding size $d = 20$.

As a large variance in the performance of the model can be observed on some tasks, we follow \cite{Sukhbaatar+:2015} and repeat each training $100$ times with different random initializations and select the best system based on the validation performance. On the 10k dataset, we repeat each training $30$ times due to time constraints. Concerning the models implementation, there are minor differences between the results of our implementation of \memnn and those reported in \cite{Sukhbaatar+:2015}, the overall performance is equally competitive and, in some cases, better. It should be noted that v1.1 of the dataset was used whereas in this work, we employ the latest v1.2. It is therefore deemed necessary that we present the performance results of our implementation of \memnn on the v1.2 dataset. To facilitate fair comparison, we select our implementation of \memnn as the baseline as we believe that it is indicative of the true performance of \memnn on v1.2 of the dataset.

\subsection{Results}
\label{sec:eval}

Performance results on the 20 \babi QA dataset are presented in \tabref{tbl:performance20bAbI}. 
For comparison purposes, we still present \memnn \cite{Sukhbaatar+:2015} in \tabref{tbl:performance20bAbI} but accompany it with the accuracy obtained by our implementation of the same model with the same experimental setup on v1.2 of the dataset in the column ``Our \memnn'' for both the 1k and 10k versions of the dataset. In contrast, we also list the results achieved by \gmemnn with global and hop-specific weight constraints in the \gmemnn columns.

\begin{table*}[tb]
\center
\resizebox{.92\textwidth}{!}{
\begin{tabular}{l|c|c@{\hskip 3mm}c@{\hskip 3mm}c||c|c@{\hskip 3mm}c@{\hskip 3mm}c}
\hline
\multirow{3}{*}{Task} & \multicolumn{4}{c||}{1k} & \multicolumn{4}{c}{10k} \\
\cline{2-5}\cline{6-9}
& \multirow{2}{*}{\small \memnn} & \small Our & \multicolumn{2}{c||}{\small \gmemnn} & \multirow{2}{*}{\small \memnn} & \small Our & \multicolumn{2}{c}{\small \gmemnn} \\
& & \small \memnn & {\small +global} & {\small +hop} & & \small \memnn & {\small +global} & {\small +hop} \\
\hline
1: 1 supporting fact & \bf 100.0 & \bf 100.0 & \bf 100.0 & \bf 100.0 & \bf 100.0 & \bf 100.0 & \bf 100.0 & \bf 100.0 \\
2: 2 supporting facts & 91.7 & 89.9 & 88.7 & \bf 91.9 & 99.7 & 99.7 & \bf 100.0 & \bf 100.0 \\
3: 3 supporting facts & 59.7 & 58.5 & 53.2 & \bf 61.2 & 90.7 & 89.1 & 94.7 & \bf 95.5\\
4: 2 argument relations & 97.2 & 99.0 & 99.3 & \bf 99.6 & \bf 100.0 & \bf 100.0 & \bf 100.0 & \bf 100.0 \\
5: 3 argument relations & 86.9 & 86.6 & 98.1 & \bf 99.0 & 99.4 & 99.4 & \bf 99.9 & 99.8\\
6: yes/no questions & \bf 92.4 & 92.1 & 92.0 & 91.6 & \bf 100.0 & \bf 100.0 & 96.7 & \bf 100.0\\
7: counting & 82.7 & 83.3 & \bf 83.8 & 82.2 & 96.3 & 96.8 & 96.7 & \bf 98.2\\
8: lists/sets & \bf 90.0 & 89.0 & 87.8 & 87.5 & 99.2 & 98.1 & \bf 99.9 & 99.7\\
9: simple negation & 86.8 & \bf 90.3 & 88.2 & 89.3 & 99.2 & 99.1 & \bf 100.0 & \bf 100.0\\
10: indefinite knowledge & \bf 84.9 & 84.6 & 80.1 & 83.5 & 97.6 & 98.0 & \bf 99.9 & 99.8\\
11: basic coreference & 99.1 & 99.7 & 99.8 & \bf 100.0 & \bf 100.0 & \bf 100.0 & \bf 100.0 & \bf 100.0\\
12: conjunction & 99.8 & \bf 100.0 & \bf 100.0  & \bf 100.0 & \bf 100.0 & \bf 100.0 & \bf 100.0 & \bf 100.0\\
13: compound coreference & 99.6 & \bf 100.0 & \bf 100.0 & \bf 100.0 & \bf 100.0 & \bf 100.0 & \bf 100.0 & \bf 100.0\\
14: time reasoning & 98.3 & \bf 99.6 & 98.5 & 98.8 & \bf 100.0 & \bf 100.0 & \bf 100.0 & \bf 100.0\\
15: basic deduction & \bf 100.0 & \bf 100.0 & \bf 100.0 & \bf 100.0 & \bf 100.0 & \bf 100.0 & \bf 100.0 & \bf 100.0\\
16: basic induction & 98.7 & \bf 99.9 & 99.8 & \bf 99.9 & 99.6 & \bf 100.0 & \bf 100.0 & \bf 100.0\\
17: positional reasoning & 49.0 & 48.1 & \bf 60.2 & 58.3 & 59.3 & 62.1 & 68.8 & \bf 72.2\\
18: size reasoning & 88.9 & 89.7 & \bf 91.8 & 90.8 & 93.3 & \bf 93.4 & 92.0 & 91.5\\
19: path finding & \bf 17.2 & 11.3 & 10.3 & 11.5 & 33.5 & 47.2 & 54.8 & \bf 69.0 \\
20: agent's motivation & \bf 100.0 & \bf 100.0 & \bf 100.0 & \bf 100.0 & \bf 100.0 & \bf 100.0 & \bf 100.0 & \bf 100.0\\
\hline
Average & 86.1 & 86.1 & 86.6 & \bf 87.3 & 93.4 & 94.1 & 95.2 & \textbf{96.3}\\
\hline
\end{tabular}
}
\caption{\label{tbl:performance20bAbI}
Accuracy (\%) on the 20 QA tasks for models using 1k and 10k training examples. \memnn:\cite{Sukhbaatar+:2015}. Our \memnn: our implementation of \memnn. \gmemnn +golbal: \gmemnn with global weight tying. \gmemnn +hop: \gmemnn with hop-specific weight tying. \textbf{Bold} highlights best performance. Note that in \cite{Sukhbaatar+:2015}, v1.1 of the dataset was used.}
\end{table*}

\myparagraph{\gmemnn achieves substantial improvements on task 5 and 17.} The performance of \gmemnn is greatly improved, a substantial gain of more than 10 in absolute accuracy. 

\myparagraph{Global vs.~hop-specific weight tying.} Compared with the global weight tying scheme on the weight matrices of the gating mechanism, applying weight constraints in a hop-specific fashion generates a further boost in performance consistently on both the 1k and 10k datasets.

\myparagraph{State-of-the-art performance on both the 1k and 10k dataset.} The best performing \gmemnn model achieves state-of-the-art performance, an average accuracy of 87.3 on the 1k dataset and 96.3 on the 10k variant. This is a solid improvement compared to \memnn and a step closer to the strongly supervised models described in \cite{WestonCB14}. Notice that the highest average accuracy of the original \memnn model on the 10k dataset is $95.8$. However, it was attained by a model with layer-wise weight tying, not adjacent weight tying as adopted in this work, and, more importantly, a much larger embedding size $d = 100$ (therefore not shown in \tabref{tbl:performance20bAbI}). In comparison, it is worth noting that the proposed \gmemnn model, a much smaller model with embeddings of size $20$, is capable of achieving better accuracy.

\section{Dialog bAbI Experiments}
\label{sec:experimentsdialog}

In addition to the text understanding and reasoning tasks presented in \secref{sec:experimentsqa}, we further examine the effectiveness of the proposed \gmemnn model on a collection of goal-oriented dialog tasks \cite{Bordes+:2016}. First, we briefly describe the dataset. Next, we outline the training details. Finally, experimental results are presented with analyses.

\subsection{Dataset and Data Preprocessing}

In this work, we adopt the goal-oriented dialog dataset developed by \newcite{Bordes+:2016} organized as a set of tasks. 
The tasks in this dataset can be divided into 6 categories with each group focusing on a specific objective: 
\begin{enumerate*}
  \item issuing API calls,
  \item updating API calls,
  \item displaying options,
  \item providing extra-information,
  \item conducting full dialogs (the aggregation of the first 4 tasks)
  \item Dialog State Tracking Challenge 2 corpus (\dstc).
\end{enumerate*}
The first 5 tasks are synthetically generated based on a knowledge base consisting of facts which define all the restaurants and their associated properties (7 types, such as location and price range). The generated texts are in the form of conversation between a user and a bot, each of which is designed with a clear yet different objective (all involved in a restaurant reservation scenario). This dataset essentially tests the capacity of end-to-end dialog systems to conduct dialog with various goals. Each dialog starts with a user request with subsequent alternating user-bot utterances and it is the duty of a model to understand the intention of the user and respond accordingly. In order to test the capability of a system to cope with entities not appearing in the training set, a different set of test sets, named out-of-vocabulary (OOV) test sets, are constructed separately. In addition, a supplementary dataset, task 6, is provided with real human-bot conversations, also in the restaurant domain, which is derived from the second Dialog State Tracking Challenge \cite{Henderson+:2014}. It is important to notice that the answers in this dataset may no longer be a single word but can be comprised of multiple ones.

\subsection{Training Details}
\label{sec:dialogtrainingdetails}

At a certain given time $t$, a memory-based model takes the sequence of utterances $c^{u}_{1}, c^{r}_{1}, c^{u}_{2}, c^{r}_{2}, \ldots, c^{u}_{t-1}, c^{r}_{t-1}$ (alternating between the user $c^{u}_{i}$ and the system response $c^{r}_{i}$) as the stories and $c^{u}_{t}$ as the question. The goal of the model is to predict the response $c^{r}_{t}$.

As answers may be composed of multiple words, following \cite{Bordes+:2016}, we replace the final prediction step in \equref{equ:finalpred} with:
\begin{equation*}
\hat{\vec{a}} = \softmax(\vec[\top]{u}\mat{W^{'}}\Phi(\vec{y}_{1}),\ldots,\vec[\top]{u}\mat{W^{'}}\Phi(\vec{y}_{|C|}))
\end{equation*}
where $\mat{W^{'}} \in \R^{d \times |V|}$ is the weight parameter matrix for the model to learn, $\vec{u} = \vec[K]{o} + \vec[K]{u}$ ($K$ is the total number of hops), $\vec{y}_{i}$ is the $i^{\textrm{th}}$ response in the candidate set $C$ such that $\vec{y}_{i} \in C$, $|C|$ the size of the candidate set, and $\Phi(\cdot)$ a function which maps the input text into a bag of dimension $|V|$.

As in \cite{Bordes+:2016}, we extend $\Phi$ by several key additional features. First, two features marking the identity of the speaker of a particular utterance (user or model) are added to each of the memory slots. Second, we expand the feature representation function $\Phi$ of candidate responses with 7 additional features, each, focusing on one of the 7 properties associated with any restaurants, indicating whether there are any exact matches between words occurring in the candidate and those in the question or memory. These 7 features are referred to as the \textit{match} features.

Apart from the modifications described above, we carry out the experiments using the same experimental setup described in \secref{sec:trainingdetails}. We also constrain ourselves to the hop-specific weight tying scheme in all our experiments since \gmemnn benefits more from it than global weight tying as shown in \secref{sec:eval}. As in \cite{Sukhbaatar+:2015}, since the memory-based models are sensitive to parameter initialization, we repeat each training $10$ times and choose the best system based on the performance on the validation set.

\subsection{Results}

Performance results on the \dialogbabi dataset are shown in \tabref{tbl:performancedialog}, measured using both per-response accuracy and per-dialog accuracy (given in parentheses). While per-response accuracy calculates the percentage of correct responses, per-dialog accuracy, where a dialog is considered to be correct if and only if every response within it is correct, counts the percentage of correct dialogs. Task 1-5 are presented in the upper half of the table while the same tasks in the OOV setting are in the lower half with dialog state tracking task as task 6 at the bottom. We choose \cite{Bordes+:2016} as the baseline which achieves the current state of the art on these tasks.

\begin{table*}[tb]
\center
\resizebox{.93\textwidth}{!}{
\begin{tabular}{l|c@{\hskip 0.5mm}cc@{\hskip 0.5mm}c||c@{\hskip 0.5mm}cc@{\hskip 0.5mm}c}
\hline
\multirow{2}{*}{Task} & \multicolumn{2}{c}{\multirow{2}{*}{\memnn}} & \multicolumn{2}{c||}{\multirow{2}{*}{\gmemnn}} & \multicolumn{2}{c}{\memnn} & \multicolumn{2}{c}{\gmemnn} \\
 & & & & & \multicolumn{2}{c}{+match} & \multicolumn{2}{c}{+match} \\
\hline
T1: Issuing API calls & 99.9 & (99.6) & \bf 100.0 & (100.0) & \bf 100.0 & (100.0) & \bf 100.0 & (100.0)\\
T2: Updating API calls & \bf 100.0 & (100.0) & \bf 100.0 & (100.0) & 98.3 & (83.9) & \bf 100.0 & (100.0)\\
T3: Displaying options & \bf 74.9 & (2.0) & \bf 74.9 & (0.0) & \bf 74.9 & (0.0) & \bf 74.9 & (0.0)\\
T4: Providing information & 59.5 & (3.0) & 57.2 & (0.0) & \bf 100.0 & (100.0) & \bf 100.0 & (100.0)\\
T5: Full dialogs & 96.1 & (49.4) & 96.3 & (52.5) & 93.4 & (19.7) & \bf 98.0 & (72.5)\\
\hline
Average & 86.1 & (50.8) & 85.7 & (50.5) & 93.3 & (60.7) & \bf 94.6 & (74.5) \\
\hline
\hline
T1 (OOV): Issuing API calls & 72.3 & (0.0) & 82.4 & (0.0) & 96.5 & (82.7) & \bf 100.0 & (100.0)\\
T2 (OOV): Updating API calls & 78.9 & (0.0) & 78.9 & (0.0) & \bf 94.5 & (48.4) & 94.2 & (47.1)\\
T3 (OOV): Displaying options & 74.4 & (0.0) & \bf 75.3 & (0.0) & 75.2 & (0.0) & 75.1 & (0.0)\\
T4 (OOV): Providing information & 57.6 & (0.0) & 57.0 & (0.0) & \bf 100.0 & (100.0) & \bf 100.0 & (100.0)\\
T5 (OOV): Full dialogs & 65.5 & (0.0) & 66.7 & (0.0) & 77.7 & (0.0) & \bf 79.4 & (0.0)\\
\hline
Average & 69.7 & (0.0) & 72.1 & (0.0) & 88.8 & (46.2) & \bf 89.7 & (49.4) \\
\hline
\hline
T6: Dialog state tracking 2 & 41.1 & (0.0) & 47.4 & (1.4) & 41.0 & (0.0) & \bf 48.7 & (1.4)\\
\hline
\end{tabular}
}
\caption{\label{tbl:performancedialog}
Per-response accuracy and per-dialog accuracy (in parentheses) on the \dialogbabi tasks. \memnn: \cite{Bordes+:2016}. +match indicates the use of the match features in \secref{sec:dialogtrainingdetails}.}
\end{table*}

\myparagraph{\gmemnn with the match features sets a new state of the art on most of the tasks.} Other than on task T2 (OOV) and T3 (OOV), \gmemnn with the match features scores the best per-response and per-dialog accuracy. Even on T2 (OOV) and T3 (OOV), the model generates rather competitive results and remains within 0.3\% of the best performance. Overall, the best average per-response accuracy in both the OOV and non-OOV categories is attained by \gmemnn.

\myparagraph{\gmemnn with the match features significantly improves per-dialog accuracy on T5.} A breakthrough in per-dialog accuracy on T5 from less than 20\% to over 70\%.

\myparagraph{\gmemnn succeeds in improving the performance on the more practical task T6.} With or without the match features, \gmemnn achieves a substantial boost in per-response accuracy on T6. Given that T6 is derived from a dataset based on real human-bot conversations, not synthetically generated, the performance gain, although far from perfect, highlights the effectiveness of \gmemnn in practical scenarios and constitutes an encouraging starting point towards end-to-end dialog system learning.

\myparagraph{The effectiveness of \gmemnn is more pronounced on the more challenging tasks.} The performance gains on T5, T5 (OOV) and T6, compared with  the rest of the tasks, are more pronounced. Regarding the performance of \memnn, these tasks are relatively more challenging than the rest, suggesting that the adaptive gating mechanism in \gmemnn is capable of managing complex information flow while doing little damage on easier tasks.

\section{Visualization and Analysis}

\begin{figure}[tb]
	\center
	\includegraphics[width=\columnwidth]{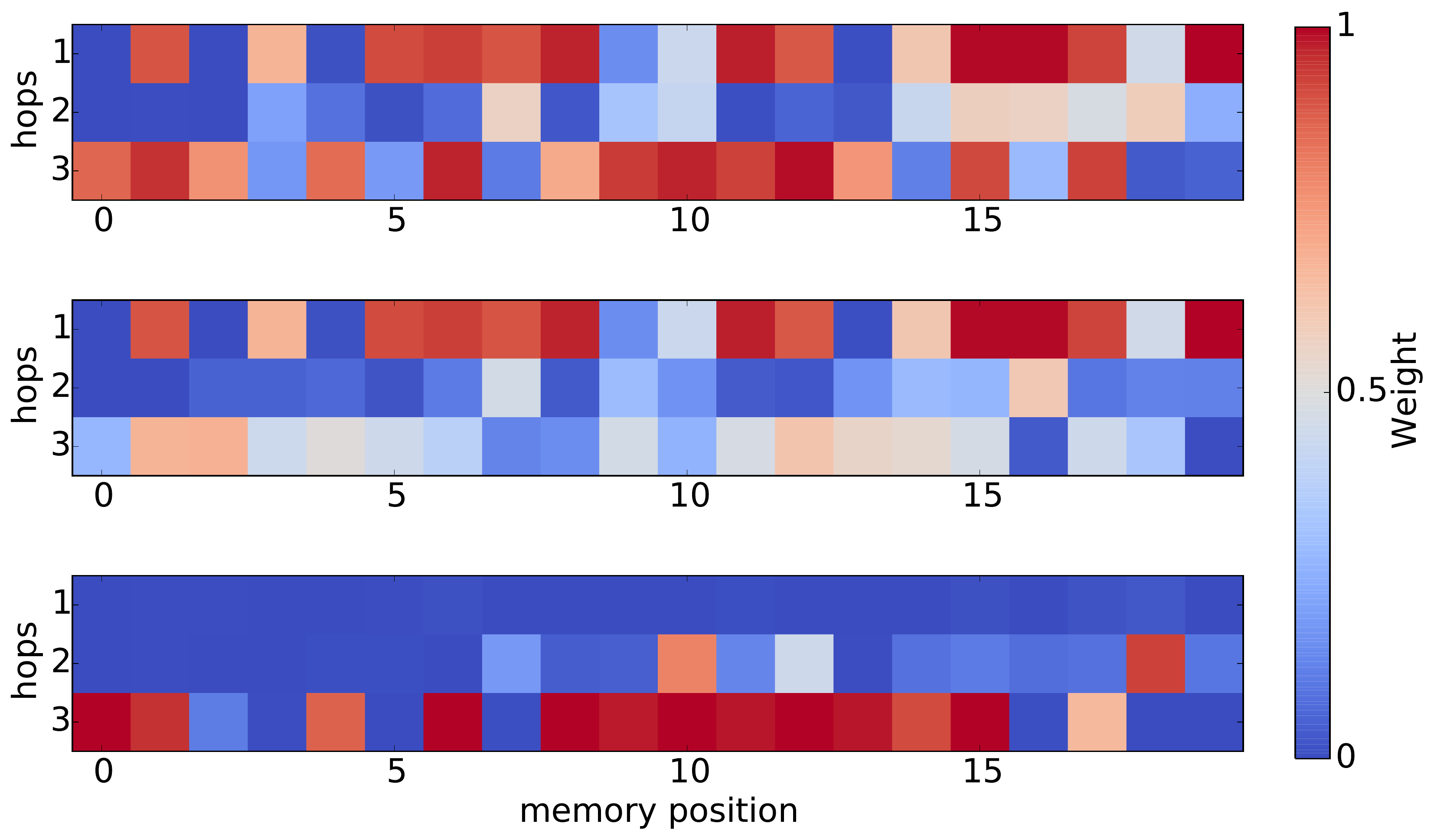}
	\caption{\label{fig:gatevalues}3 most frequently observed gate value $\textrm{T}^{k}(\vec[k]{u})$ patterns on T6 of the \dialogbabi dataset}
\end{figure}

\begin{figure}[tb]
	\center
	\includegraphics[width=\columnwidth]{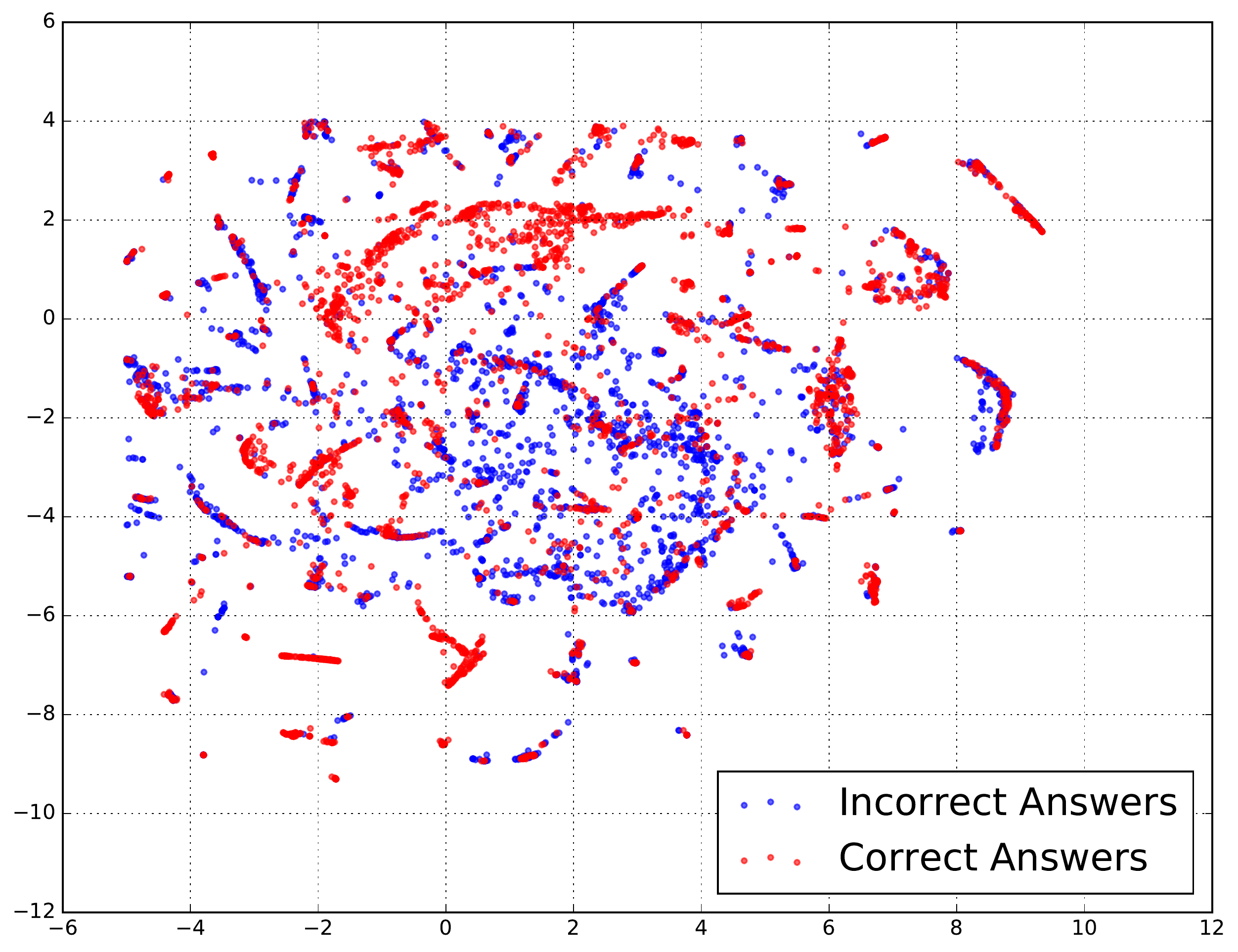}
	\caption{\label{fig:tsne}t-SNE scatter plot of the flattened gate values}
\end{figure}

In addition to the quantitative results, we further look into the memory regulation mechanism learned by the \gmemnn model. \figref{fig:gatevalues} presents the three most frequently observed patterns of the $\textrm{T}^{k}(\vec[k]{u})$ vectors for each of the $3$ hops in a model trained on T6 of the \dialogbabi dataset with an embedding dimension of $20$. Each row corresponds to the gate values at a specific hop whereas each column represents a given embedding dimension. The pattern on the top indicates that the model tends to only access memory in the first and third hop. In contrast, the middle and bottom patterns only focus on the memory in either the first or last hop respectively. \figref{fig:tsne} is a t-SNE projection \cite{Maaten08} of the flattened $[\textrm{T}^{1}(\vec[1]{u});\textrm{T}^{2}(\vec[2]{u});\textrm{T}^{3}(\vec[3]{u})]$ vectors obtained on the test set of the same dialog task with points corresponding to the correct and incorrect responses in red and blue respectively. Despite the relative uniform distribution of the wrong answer points, the correct ones tend to form clusters that suggest the frequently observed behavior of a successful inference. 
Lastly, \tabref{tbl:memnnvsgmemnn} shows the comparison of the attention shifting process between \memnn and \gmemnn on a story on \babi task 5 (3 argument relations). Not only does \gmemnn manage to focus more accurately on the supporting fact than \memnn, it has also learned to rely less in this case on hop 1 and 2 by assigning smaller transform gate values. 
In contrast, \memnn carries false and misguiding information (caused by the distracting attention mechanism) accumulated from the previous hops, which eventually led to the wrong prediction of the answer.

\begin{table*}[tb]
\centering
\resizebox{0.9\textwidth}{!}{
\begin{tabular}{|lc|ccc|ccc|}
\hline
\multirow{2}{*}{Story} & \multirow{2}{*}{Support} & \multicolumn{3}{c|}{\memnn} & \multicolumn{3}{c|}{\gmemnn} \\
& & Hop 1 & Hop 2 & Hop 3 & Hop 1 & Hop 2 & Hop 3 \\
\hline
Fred took the football there. & & \cellcolor{red!4}0.05 & \cellcolor{red!10}0.10 & \cellcolor{red!6}0.07 & \cellcolor{red!6}0.06 & \cellcolor{red!0}0.00 & \cellcolor{red!0}0.00\\
Fred journeyed to the hallway. & & \cellcolor{red!45}0.45 & \cellcolor{red!9}0.09 & \cellcolor{red!0}0.01 & \cellcolor{red!0}0.00 & \cellcolor{red!0}0.00 & \cellcolor{red!0}0.00\\
Fred passed the football to Mary. & yes & \cellcolor{red!10}0.10 & \cellcolor{red!64}0.64 & \cellcolor{red!92}0.93 & \cellcolor{red!29}0.29 & \cellcolor{red!99}1.00 & \cellcolor{red!99}1.00\\
Mary dropped the football. & & \cellcolor{red!40}0.40 & \cellcolor{red!17}0.17 & \cellcolor{red!0}0.00 & \cellcolor{red!64}0.64 & \cellcolor{red!0}0.00 & \cellcolor{red!0}0.00\\
\hline
\multicolumn{2}{|l|}{Avg. transform gate cell values, $\sum_i\textrm{T}^{k}(\vec[k]{u})_i/d$} & N/A & N/A & N/A & 0.22 & 0.23 & 0.45 \\
\hline
\multicolumn{8}{|l|}{\specialcell{Question: Who gave the football? Answer: Fred, \memnn: {\color{red} Mary}, \gmemnn: Fred}}\\
\hline
\end{tabular}
}
\caption{\label{tbl:memnnvsgmemnn}\memnn vs.~\gmemnn - \babi dataset - Task 5 - 3 argument relations}
\end{table*}

\section{Related Reading Tasks} 

Apart from the datasets adopted in our experiments, the CNN/Daily Mail \cite{Hermann+:2015} has been used for the task of machine reading formalized as a problem of text extraction from a source conditioned on a given question. However, as pointed out in \cite{Chen+:2016}, this dataset not only is noisy but also requires little reasoning and inference, which is evidenced by a manual analysis of a randomly selected subset of the questions, showing that only 2\% of the examples call for multi-sentence inference. 
\newcite{Richardson+:2013} constructed an open-domain reading comprehension task, named MCTest. Although this corpus demands various reasoning capabilities from multiple sentences, its rather limited size (660 paragraphs, each associated with 4 questions) renders training statistical models infeasible \cite{Chen+:2016}. Children's Book Test (CBT) \cite{Hill+:2015} was designed to measure the ability of models to exploit a wide range of linguistic context. Despite the claim in \cite{Sukhbaatar+:2015} that increasing the number of hops is crucial for the performance improvements on some tasks, which can be seen as enabling \memnn to accommodate more supporting facts, making such performance boost particularly more pronounced on those tasks requiring complex reasoning, \newcite{Hill+:2015} admittedly reported little improvement in performance by stacking more hops and chose a single-hop \memnn. This suggests that the necessity of multi-sentence based reasoning on this dataset is not mandatory. In the future, we plan to investigate into larger dialog datasets such as \cite{Lowe+:2015}.

%% file: conc.tex
\section{Conclusion and Future Work}

In this paper, we have proposed and developed what is, as far as our knowledge goes, the first attempt at incorporating an iterative memory access control to an end-to-end trainable memory-enhanced neural network architecture. We showed the added value of our proposition on a set of, natural language based, state-of-the-art reasoning tasks. Then, we offered a first interpretation of the resulting capability by analyzing the attention shifting mechanism and connection short-cutting behavior of the proposed model. In future work, we will investigate the use of such mechanism in the field of language modeling and more generally on the paradigm of sequential prediction and predictive learning. Furthermore, we plan to look into the impact of this method on the recently introduced Key-Value Memory Networks \cite{Miller+:2016} on larger and semi-structured corpus.

%% file: main.bbl
\begin{thebibliography}{}

\bibitem[\protect\citename{Bishop}1995]{Bishop1995}
Christopher~M. Bishop.
\newblock 1995.
\newblock {\em Neural Networks for Pattern Recognition}.
\newblock Oxford University Press.

\bibitem[\protect\citename{Bordes and Weston}2016]{Bordes+:2016}
Antoine Bordes and Jason Weston.
\newblock 2016.
\newblock Learning end-to-end goal-oriented dialog.
\newblock {\em arXiv preprint arXiv:1605.07683}.

\bibitem[\protect\citename{Chen \bgroup et al.\egroup }2016]{Chen+:2016}
Danqi Chen, Jason Bolton, and Christopher~D Manning.
\newblock 2016.
\newblock A thorough examination of the cnn/daily mail reading comprehension
  task.
\newblock In {\em Proceedings of the 54th Annual Meeting of the Association for
  Computational Linguistics (ACL 2016)}, pages 2358--2367, Berlin, Germany.

\bibitem[\protect\citename{Glorot and Bengio}2010]{GlorotB10}
Xavier Glorot and Yoshua Bengio.
\newblock 2010.
\newblock Understanding the difficulty of training deep feedforward neural
  networks.
\newblock In {\em Proceedings of the 13th International Conference on
  Artificial Intelligence and Statistics (AISTATS 2010)}, pages 249--256,
  Sardinia, Italy.

\bibitem[\protect\citename{He \bgroup et al.\egroup }2015]{HeZRS15}
Kaiming He, Xiangyu Zhang, Shaoqing Ren, and Jian Sun.
\newblock 2015.
\newblock Delving deep into rectifiers: Surpassing human-level performance on
  imagenet classification.
\newblock In {\em Proceedings of the IEEE International Conference on Computer
  Vision (ICCV 2015)}, pages 1026--1034, Santiago, Chile.

\bibitem[\protect\citename{He \bgroup et al.\egroup }2016]{HeZRS15a}
Kaiming He, Xiangyu Zhang, Shaoqing Ren, and Jian Sun.
\newblock 2016.
\newblock Deep residual learning for image recognition.
\newblock In {\em the 29th IEEE Conference on Computer Vision and Pattern
  Recognition (CVPR 2016)}, Las Vegas, USA.

\bibitem[\protect\citename{Henderson \bgroup et al.\egroup
  }2014]{Henderson+:2014}
Matthew Henderson, Blaise Thomson, and Jason~D. Williams.
\newblock 2014.
\newblock The second dialog state tracking challenge.
\newblock In {\em Proceedings of the 15th Annual Meeting of the Special
  Interest Group on Discourse and Dialogue (SIGDIAL 2014)}, pages 263--272,
  Philadelphia, USA.

\bibitem[\protect\citename{Hermann \bgroup et al.\egroup }2015]{Hermann+:2015}
Karl~Moritz Hermann, Tomas Kocisky, Edward Grefenstette, Lasse Espeholt, Will
  Kay, Mustafa Suleyman, and Phil Blunsom.
\newblock 2015.
\newblock Teaching machines to read and comprehend.
\newblock In {\em Proceedings of Advances in Neural Information Processing
  Systems (NIPS 2015)}, pages 1684--1692, Barcelona, Spain.

\bibitem[\protect\citename{Hill \bgroup et al.\egroup }2015]{Hill+:2015}
Felix Hill, Antoine Bordes, Sumit Chopra, and Jason Weston.
\newblock 2015.
\newblock The goldilocks principle: Reading children's books with explicit
  memory representations.
\newblock In {\em Proceedings of the 4th International Conference on Learning
  Representations (ICLR 2016)}, San Juan, Puerto Rico.

\bibitem[\protect\citename{Kumar \bgroup et al.\egroup }2016]{KumarIOIBGZPS16}
Ankit Kumar, Ozan Irsoy, Jonathan Su, James Bradbury, Robert English, Brian
  Pierce, Peter Ondruska, Ishaan Gulrajani, and Richard Socher.
\newblock 2016.
\newblock Ask me anything: Dynamic memory networks for natural language
  processing.
\newblock In {\em Proceedings of the 33rd International Conference on Machine
  Learning (ICML 2016)}, New York, USA.

\bibitem[\protect\citename{LeCun \bgroup et al.\egroup }1998]{LeCun98}
Yann LeCun, Leon Bottou, Genevieve~B. Orr, and Klaus~Robert M{\"u}ller.
\newblock 1998.
\newblock Efficient backprop.
\newblock {\em Neural Networks: Tricks of the Trade}, pages 9--50.

\bibitem[\protect\citename{Lowe \bgroup et al.\egroup }2015]{Lowe+:2015}
Ryan Lowe, Nissan Pow, Iulian Serban, and Joelle Pineau.
\newblock 2015.
\newblock The ubuntu dialogue corpus: A large dataset for research in
  unstructured multi-turn dialogue systems.
\newblock In {\em Proceedings of the 16th Annual SIGdial Meeting on Discourse
  and Dialogue (SIGDIAL 2015)}, Prague, Czech Republic.

\bibitem[\protect\citename{Maaten and Hinton}2008]{Maaten08}
Laurens van~der Maaten and Geoffrey Hinton.
\newblock 2008.
\newblock Visualizing data using t-{SNE}.
\newblock {\em Journal of Machine Learning Research}, 9:2579--2605.

\bibitem[\protect\citename{Miller \bgroup et al.\egroup }2016]{Miller+:2016}
Alexander Miller, Adam Fisch, Jesse Dodge, Amir-Hossein Karimi, Antoine Bordes,
  and Jason Weston.
\newblock 2016.
\newblock Key-value memory networks for directly reading documents.
\newblock In {\em Proceedings of the 2016 Conference on Empirical Methods in
  Natural Language Processing (EMNLP 2016)}, Austin, USA.

\bibitem[\protect\citename{Richardson \bgroup et al.\egroup
  }2013]{Richardson+:2013}
Matthew Richardson, Christopher~J.C. Burges, and Erin Renshaw.
\newblock 2013.
\newblock {MCTest}: A challenge dataset for the open-domain machine
  comprehension of text.
\newblock In {\em Proceedings of the 2013 Conference on Empirical Methods in
  Natural Language Processing (EMNLP 2013)}, pages 193--203, Seattle, USA.

\bibitem[\protect\citename{Ripley}2007]{ripley96pattern}
Brian~D. Ripley.
\newblock 2007.
\newblock {\em Pattern recognition and neural networks}.
\newblock Cambridge university press.

\bibitem[\protect\citename{Saxe \bgroup et al.\egroup }2014]{SaxeMG13}
Andrew~M Saxe, James~L McClelland, and Surya Ganguli.
\newblock 2014.
\newblock Exact solutions to the nonlinear dynamics of learning in deep linear
  neural networks.
\newblock In {\em Proceedings of the 2nd International Conference on Learning
  Representations (ICLR 2014)}, Banff, Canada.

\bibitem[\protect\citename{Srivastava \bgroup et al.\egroup
  }2015a]{SrivastavaGS15}
Rupesh~Kumar Srivastava, Klaus Greff, and J{\"u}rgen Schmidhuber.
\newblock 2015a.
\newblock Highway networks.
\newblock {\em arXiv preprint arXiv:1505.00387}.

\bibitem[\protect\citename{Srivastava \bgroup et al.\egroup
  }2015b]{Srivastava+:2015}
Rupesh~Kumar Srivastava, Klaus Greff, and J{\"u}rgen Schmidhuber.
\newblock 2015b.
\newblock Training very deep networks.
\newblock In {\em Proceedings of Advances in Neural Information Processing
  Systems (NIPS 2015)}, pages 2377--2385, Montr\'{e}al, Canada.

\bibitem[\protect\citename{Sukhbaatar \bgroup et al.\egroup
  }2015]{Sukhbaatar+:2015}
Sainbayar Sukhbaatar, Arthur Szlam, Jason Weston, and Rob Fergus.
\newblock 2015.
\newblock End-to-end memory networks.
\newblock In {\em Proceedings of Advances in Neural Information Processing
  Systems (NIPS 2015)}, pages 2440--2448, Montr\'{e}al, Canada.

\bibitem[\protect\citename{Weston \bgroup et al.\egroup }2015]{WestonCB14}
Jason Weston, Sumit Chopra, and Antoine Bordes.
\newblock 2015.
\newblock Memory networks.
\newblock In {\em Proceedings of the 3rd International Conference on Learning
  Representations (ICLR 2015)}, San Diego, USA.

\bibitem[\protect\citename{Weston \bgroup et al.\egroup }2016]{WestonBCM15}
Jason Weston, Antoine Bordes, Sumit Chopra, Alexander~M Rush, Bart van
  Merri{\"e}nboer, Armand Joulin, and Tomas Mikolov.
\newblock 2016.
\newblock Towards {AI}-complete question answering: {A} set of prerequisite toy
  tasks.
\newblock In {\em Proceedings of the 4th International Conference on Learning
  Representations (ICLR 2016)}, San Juan, Puerto Rico.

\bibitem[\protect\citename{Xiong \bgroup et al.\egroup }2016]{XiongMS16}
Caiming Xiong, Stephen Merity, and Richard Socher.
\newblock 2016.
\newblock Dynamic memory networks for visual and textual question answering.
\newblock In {\em Proceedings of the 33rd International Conference on Machine
  Learning (ICML 2016)}, pages 2397--2406, New York, USA.

\end{thebibliography}
